# Codebook Reduction and Saturation: Novel observations on Inductive Thematic Saturation for Large Language Models and initial coding in Thematic Analysis

Stefano De Paoli – Abertay University – s.depaoli@abertay.ac.uk

Walter Stan Mathis – Yale Medical School - walter.mathis@yale.edu

*[this is a draft it may contain some errors, please get in touch with the authors in case you spot any issue, and we will correct them]*

## Abstract

This paper reflects on the process of performing Thematic Analysis with Large Language Models (LLMs). Specifically, the paper deals with the problem of analytical saturation of initial codes, as produced by LLMs. Thematic Analysis is a well-established qualitative analysis method composed of interlinked phases. A key phase is the initial coding, where the analysts assign labels to discrete components of a dataset. Saturation is a way to measure the validity of a qualitative analysis and relates to the recurrence and repetition of initial codes. In the paper we reflect on how well LLMs achieve analytical saturation and propose also a novel technique to measure Inductive Thematic Saturation (ITS). This novel technique leverages a programming framework called DSPy. The proposed novel approach allows a precise measurement of ITS.

## Introduction

Large Language Models (LLMs) are a type of Generative Artificial Intelligence systems, which have significant capacity to manipulate language and produce high quality textual outputs. Known examples include the GPT models (e.g. GPT4o), Llama-3, DeepSeek or Google Gemini. LLMs have recently started to be used for performing qualitative analysis on textual data, such as semi-structured interviews. Initial studies demonstrate promising results. However, methodological research is still a high priority in this nascent sub-field to use novel techniques in sound ways. Thematic Analysis (TA) is one of the approaches currently being explored (see e.g. De Paoli, 2024,

Mathis et al, 2024 or Drápal et al., 2023; Chen et al., 2024 for some examples), with existing procedures following, in broad strokes, the six phases approach to TA formulated by Braun and Clarke (2006). These six phases include: (1) familiarisation with the data (e.g. a dataset of qualitative interviews); (2) initial coding of the data; (3) identification of themes from the codes; (4) revision of themes; (5) renaming and summarising of themes; (6) write-up of the results.

Alongside publications seeking to establish the procedures for performing TA with LLMs, authors occasionally also reflected on aspects associated with the validity of the methods. Most existing publications have focused on either inter-coder reliability (between a human analyst and an LLM) or in general with other forms of comparisons between LLMs and human analysts (see e.g. Long et al., 2024; Törnberg, 2024). Initial work has also been proposed (see De Paoli and Mathis, 2024) to assess LLMs capacity to provide analytical saturation during initial coding. Saturation is often regarded as an important metric for measuring the validity of qualitative analysis. It refers to the point at which no new codes or insights are emerging from the data. Saturation is achieved when additional coding of data does not yield new categories, or concepts, ensuring that the analysis is comprehensive and well-rounded. Inductive Thematic Saturation is a specific type of saturation which measures how well the initial codes saturate the data (Saunders et al., 2018), which we abbreviate as ITS. This paper's aim is to extend the authors' previous work on measuring LLMs ITS during the initial coding phase of TA.

Initial coding is a key step in qualitative analysis, encompassing the breakdown of textual data in small but relevant components, each of which is associated with a label (the code), which captures the meaning of the text (Saldaña, 2021). The result is a list of initial codes, which can be used to create second order categories (like themes, or axial coding). Our approach for performing initial coding of TA sees the LLMs producing a set of codes for each interview in a dataset. Each interview is coded independently with the LLM having no memory of previous interviews or codes. This 'independent initial coding' results in codes being repeated across the data analysis. The list of codes is then refined by another LLM pass to eliminate duplicates or those with essentially the same meaning, ensuring a comprehensive and efficient set of **unique codes.** Measuring saturation in this case relies therefore on removing and merging duplicate initial codes and then computing the ratio relating the overall number of unique codes (no duplicates) and of the **total codes** (including duplicates). This results in a synthetic ITS metric, between 0 and 1. Details about the metric will be presented later in the paper. In previous work we then measured saturation using as examples two datasets of interviews (n=9 and n=39) (De Paoli and Mathis, 2024). Further work, however, is needed to improve the process for measuring LLMs' ITS. Consequently, in the following pages we will address the following two objectives:

1. **Secondary Objective**: Explore further the emerging properties of ITS for LLMs, observing dynamics related with the repetition of the creation of the initial codes and the codebook reduction with the same sequence of interviews, but also with different sequences.
2. **Primary Objective**: Based on the observations from the secondary objective, propose a novel procedure for the reduction of duplicates from the total codebook, which offers better quality results (albeit at a higher cost), based on the programming framework called DSPy. This procedure also mitigates some of the issues related to the emerging properties of saturation for LLMs.

**Literature review**

Saturation is often considered an important element for evaluating the validity of a qualitative analysis, even though agreement on its value varies. Traditionally the idea is attributed to Glaser and Strauss (1967, p. 1) and their work on Grounded Theory, where in an essential definition it means "no additional data are being found whereby the sociologist can develop properties of the category". Often this is interpreted in relation to the sample of data used in the analysis. If new categories (or codes) still emerge from the analysis, this suggests that saturation has not been reached, and the researcher should collect more data up to the point that analysis of a new data element (e.g. an interview) does not produce any new category. In practice, however, the concept is more multidimensional than this.

Authors have noted the complexity of defining saturation and the problem of applying the notion across all sorts of qualitative studies (O'Reilly and Parker, 2012). This observation is loosely similar to comments offered by Braun and Clarke (2021), who encouraged scholars performing reflexive TA to use judgements during interpretation, rather than operationalised metrics. Whilst the notion of saturation has a certain currency in qualitative studies, it is also a concept not always applicable, not always considered a measure of validity, and often difficult to define. Sebele-Mpofu (2018) discussed through an extensive review of literature how the definition of saturation is a 'controversy' and notes disagreements across definitions as well as the applicability of the idea in practical terms. This echoes an earlier input by Fusch and Ness (2015) who observed that "there is no one-size-fits-all method to reach data saturation; moreover, more is not necessarily better than less and vice versa." (p. 1413). This quote signals that while the notion of saturation is important it cannot be applied mechanically. There is no guarantee, for example, that gathering more data is always better for analysis.

More interesting than the focus on the controversies are the observations that saturation may be seen as a multidimensional problem. Certain forms of saturation may apply to certain studies rather than others. A relevant typology was proposed by Saunders et al. (2018) which distinguished: (1) theoretical saturation, (2) inductive thematic saturation, (3) a priori thematic saturation and (4) data saturation. Theoretical saturation reflects the definition by Glaser and Strauss (1967). A priori thematic saturation reflects how well codes and categories are illustrated by the data and the authors connect it with the problem of sampling. Data saturation is related to the need to gather more data to satisfy the analysis. Inductive thematic saturation, lastly, is seen as having analytical properties relating to whether new codes keep emerging from the data. This last is less related to sampling or the need to collect data than the other three, as it reflects how well the categories saturate with the data already collected. This typology was further elaborated by Sebele-Mpofu (2018), considering the following forms: (a) theoretical saturation, (b) thematic saturation, (c) data saturation and (d) meaning saturation. The author collapsed the saturations (1) and (3) in Saunder et al. in the theoretical case (a) and added a new type relating to 'meaning' which is said to encompass the whole qualitative research process. As we noted previously (see De Paoli and Mathis, 2024), when working with LLMs on TA, the kind of saturation which appears relevant is the analytical type, since an LLM does not have any decision in the sampling of data, or in the data collection. The case (2) in Saunders et al. and the equivalent case (b) in Sebele-Mpofu, is the type of saturation to consider when assessing LLMs capacity to perform a valid TA.

Studies have been produced to identify a good or optimal number of interviews for reaching saturation. Guidelines, methodologies and empirical observations have been proposed. Whilst this is not immediately relevant for our work here, as we do not deal with sampling or data collection saturation, we would like to note the influential paper by Guest et al. who observed (in a sixty interviews analysis), that the large majority of their codes (92%) was observed by the 12th interview. This study puts a number on the size of sampling needed for reaching Theoretical Saturation. Followers have further developed techniques to determine the tag number of interviews for saturation, one example is Rowland et al. (2015) which likened saturation to the principle of diminishing returns and proposed a mathematical formulation for calculating the number of interviews for reaching saturation. Lowe et al. (2018) proposed a statistical-mathematical model to calculate saturation based on assessing the production of themes and predicting a saturation index based on an assessment of the potential number of all possible themes. This proposition, in our opinion, appears quite complex to calculate and the underpinning mathematical model does add complexity, whereas other authors presented simpler approaches for a similar outcome. For example, Guest et al. (2020) proposed a method to measure saturation by assessing ratios of Base Size (an initial set of interviews), Run Length (additional interviews, e.g. n=2), and New Information Threshold which is a sort of confidence level that saturation has occurred. This gives a metric for saturation which puts in relation the number of e.g. themes emerging from the base (as denominator) and the ones emerging from the Run Length. This metric has some relevant features such as being easy to calculate, and it can be applied both during the data collection and analysis as well as when the analysis is completed. However, this still is a metric for measuring data and sampling related saturation, and not Inductive Thematic Saturation.

An interesting study, which we will partly replicate later, was conducted by Costantinou et al. (2017). The authors proposed an empirical method for assessing saturation when performing TA. This is based on comparing interviews coding with each other but also on performing the analysis with interviews in different order/sequences. This second aspect of the work is relevant for us. First, the authors do consider saturation not in terms of sampling but rather that "what is saturated is not the raw data but rather the categories or themes" (p. 583). Their method proposes to compare the initial coding of each interview to one another (in order of analysis) to identify the point (the interview number) at which saturation occurs in the form of no new codes emerging from the analysis. They performed the process with 4 different sequences of interviews, using a dataset of 12 interviews. The authors performed the process starting from interview 1 to interview 12 (Sequence 1). Then in reverse order (from 12 to 1, Sequence 2) and then with two other sequences (Sequence 3: 6, 10, 9, 4, 12, 11, 7, 8, 1, 2, 3, 5 and Sequence 4: 4, 2, 1, 11, 10, 7, 12, 9, 6, 3, 5, 8).

They showed that with the first sequence saturation was reached with the 5th interview. The last sequence instead allowed them to reach saturation after the 8th interview. Whereas the two remaining sequences allowed them to reach saturation after the 6th and 7th interview. This shows that the order in which interviews are analysed may have an effect on the saturation point. They call this the order-induced error.

The literature above gives us an overview of complexity of the notion, the evidence that saturation has multiple facets, and the effort of the scientific community to operationalise the concept. The underlying idea for our own work then is to reflect on ITS when LLMs are used for performing TA and initial coding.

**LLM saturation**

Since, in the procedure currently available for TA, an LLM analyses each interview independently, the initial codes a model generates for each interview are not influenced by previous ones. This results in some redundancy across different interviews, a scenario that is less common when human analysts conduct the coding. Researchers using qualitative analysis software can reuse codes from earlier interviews to label new data that conveys similar meanings. However, LLMs process each interview in isolation, which leads to the generation of duplicate codes. Therefore, analysts using an LLM must identify and remove these duplicates to compile a final codebook of **unique codes**, which are then used for the identification of themes. In a study aimed at testing analytical saturation with LLMs, we proposed a method to construct the codebook of unique codes in a cumulative manner. This involves comparing each set of codes generated by the LLM for each interview against a cumulative codebook of unique codes. If a code already exists (in the same or very similar form) in the **unique cumulative codebook** – UCC - (i.e., it is a duplicate), the LLM discards it. Conversely, if the code is not present (i.e., it identifies a novel aspect of data), it is added to the UCC. Initially, the UCC is identical to the codes from the first interview. In subsequent steps, as the codes from all interviews are compared, new unique codes are continuously added to the UCC (see Figure 1).

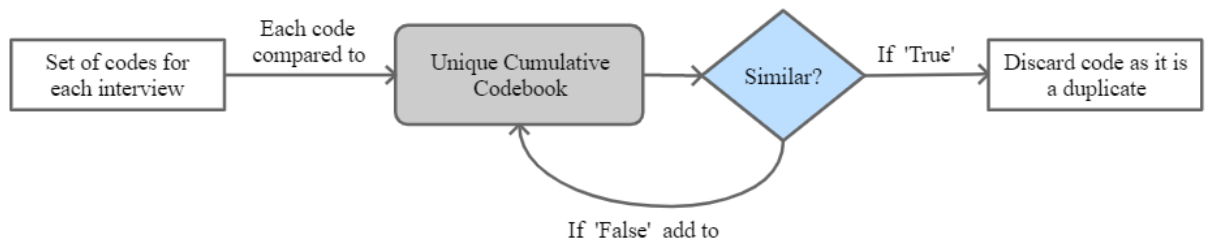

Figure 1. Reduction of duplicate codes process as presented in De Paoli and Mathis (2024)

At the end of process, we have two codebooks; (1) a **total cumulative codebook** (TCC) with all the codes, including duplicates; (2) a **unique cumulative codebook** (UCC) without duplicates (unique codes). It is this second codebook that can then be used for the identification of themes (i.e. phase 3 of TA). Moreover, removing duplicate codes allows us to obtain the figures of the total number of codes and the total unique codes for each interview point from which we can compute saturation.

In our earlier work, we showed that the sets of codes in the UCC and the TCC grow in some kind of linear fashion. The UCC expands more slowly than the TCC due to the removal of duplicates. We hypothesised that this slower growth indicates a form of saturation, and that the ratio of the total unique codes to total number of codes can serve as a measure of ITS. This ratio can be calculated using the slope ratio of the two functions, or in even simpler terms the ratio of the overall cumulative figures (unique/total). This ratio ranges from 0 (ideal case where all interviews are identical and the LLM generates the same initial codes for each interview) to 1 (where all interviews are entirely heterogenous, and the LLM generates always unique codes for each). For ITS, the ratio should not be too close to 1, indicating some degree of saturation. Figure 2 provides a simplified, fictitious illustration of this concept, showing the slower growth of the UCC compared to the TCC, on a hypothetical 16 interviews dataset. A real example will be shown later. The ratio UCC/TCC at the final interview (e.g., 16) provides the ITS metric. In the example (Figure 2), this is 67 unique codes out of 235 total codes, resulting in a ratio of 67/235, or approximately 0.29.

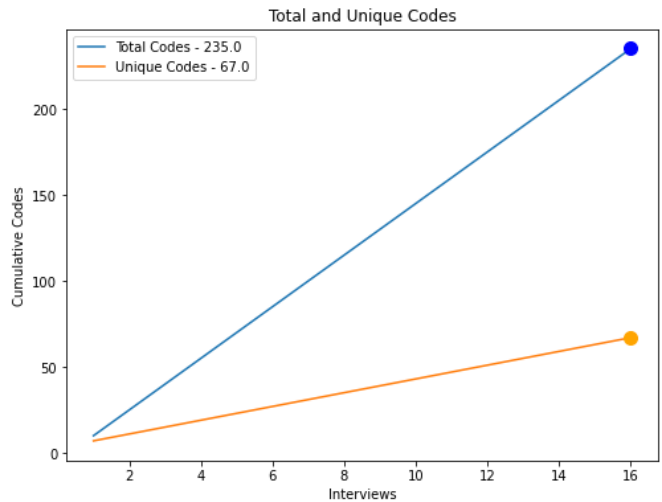

Figure 2. Simplified interpretation of saturation of initial codes for LLMs.

## Methods

The problems we seek to investigate in this paper are two-fold. The main objective is to propose an improved procedure for the reduction of the codebook generated by the LLM (from TCC to the UCC), allowing a more precise measure of the ITS metric. This main objective is however predicated on observing (and addressing) some emerging behaviours which LLMs do seem to display during the initial coding phase of TA.

**Design and Procedure**

We will perform the initial coding process on a dataset of semi-structured qualitative interviews (n=12). We will perform the process multiple times on the same sequence of interviews (from 1 to 12) and then repeat the initial coding using the additional sequences proposed by Costantinou et al. presented earlier in the literature review, and two additional sequences. The goal is to identify any variation in the LLMs response (the initial codes for each interview) caused by the probabilistic nature of the model response, and any effect of the order of analysis on saturation.

To produce the initial codes, we use a prompt tested in previous research work. A prompt is the set of instructions given to an LLM for a task to complete, triggering a response from the model. The prompt is passed to the model programmatically via python and the Application Program Interface (API) which returns a response. For doing the initial coding Prompt_1 presented in Box 1, requests the LLM (GPT3.5-Turbo-16k was used in this research, but other LLMs such as higher GPT ones, Llama or others would work too) to identify a fixed set of initial codes (up to 15) for an interview, alongside a description of the code and an exemplary quote. Each interview is passed in its entirety within the prompt in a variable (**{text}**) and then to the LLM API. The model returns a set of initial codes (with a description of the code and a representative quote) for each interview in the dataset (12 sets of initial codes), formatted as a JSON. Each JSON is then stored in separate csv files. Because each set of codes is produced independently there are duplicate codes, or codes which are very similar to one another. In the second step of the process, the codebook is reduced to a set of unique codes by comparing each interview set with a cumulative codebook of unique codes, to arrive at the end at a final set of unique codes (UCC). The process is depicted in Figure 3.

---

**PROMPT_1**

prompt = f"""

Can you assist me in the generation of initial codes to assist me with my thematic analysis. \

Identify the 15 most relevant initial codes in the text, provide a meaningful name for each code in no more than 5 words, 30 words simple description of the code, and a max 40 words quote from the participant.\

Format the response as a json file keeping names, descriptions and quotes together in the json, and keep them together in 'Codes'.

```{text}```

"""

```
    response = get_completion(prompt)
```

Box 1 – Prompt for initial coding

In this paper we will perform the codebook reduction of duplicates with two different processes. The first reproposes the process described in our previous paper (De Paoli and Mathis, 2024). In this case the codebook reduction is performed with a further prompt (see Box 2, Prompt_2) passed with python to the model API. The model is requested to compare each 'code – description' pair (contained in the variable **{value}**) to the cumulative list of unique codes, to identify if said code is already in the list or not (in which case the code is not a duplicate and will be added to the list). The UCC is built iteratively after each comparison, and at the first step the set of codes of the first interview is equal to the UCC. Unique codes are added at each iteration and the duplicates are discarded. When the last interview is compared, the process is completed. The cumulative unique and total number of codes can then be used to calculate the saturation metric as a ratio.

**PROMPT_2**

```
prompt = f"""

Then, determine if value: ```{value}``` conveys a resembling idea or meaning

to any element in the list combined_unique: {", ".join(combined_u)}.

Your response should be either a string 'true' (Similar idea or meaning) or a string 'false' (no similarity).

Format the response as a json file using the key value_in_combined_unique

"""
```

Box 2 – Prompt used for reduction of duplicates (combined_u is the list of unique codes after each iteration)

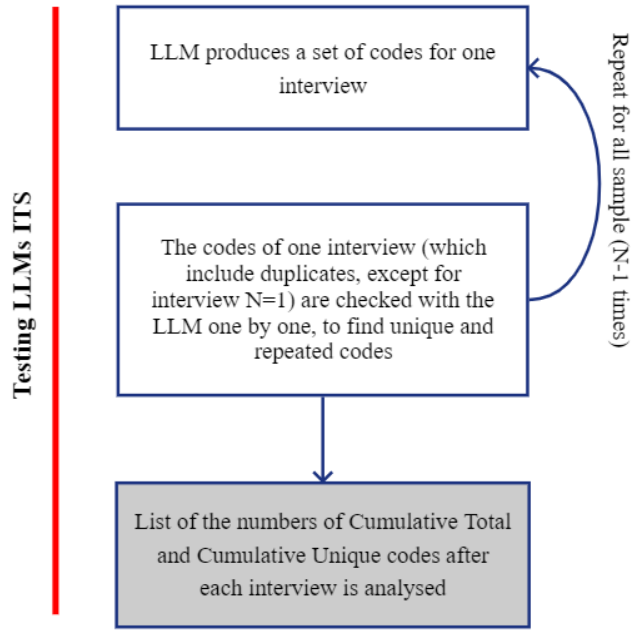

Figure 3. Process for initial coding and codebook reduction to find unique codes (from De Paoli and Mathis, 2024)

The second process we propose for removing duplicate codes uses a different way of prompting the LLM, leveraging the DSPy framework. As we will see later, in performing the initial coding procedure described above, we noticed the LLM output is not always consistent in the number of initial codes produced with Prompt_1. We attributed this to the probabilistic response of the LLM, that even with Temperature at 0 ('no creativity'), still provides responses that may be partially different from one another. Consequently, the codebook reduction with Prompt_2 could also be affected as well as the ITS metric. We started consequently looking at options to render the process more consistent.

DSPy (Katthab et al. 2023) is a framework designed for generating LLM prompts through demonstrations, leveraging iterative revision and automatic evaluation to refine and enhance the prompts. The model's prompts are built programmatically via python pipelines and examples given to the LLM, rather than through trial-and-error tests. The codebook reduction procedure appeared to us a task that could benefit from this approach. The process we designed uses a list of examples which was built from four existing open access datasets of initial coding retrieved from Zenodo and other Open Access repositories (Novelskaitė and Pučėtaitė, 2021; Antunes et al., 2019; Gilfoyle et al., 2021; Suarez, 2021), and a DSPy signature.

The signature is a DSPy concept: "a declarative specification of input/output behavior of a DSPy module. Signatures allow you to tell the LM what it needs to do, rather than specify how we should

ask the LM to do it."[1] The signature we propose for the codebook reduction is as follows (in python):

**dspy.ChainOfThought('text_1, text_2 -> meaning')**

It takes two texts, two pairs of 'code-description', instructing the LLM to check if they convey a similar meaning or not.

DSPy allows prompts to be built (via the signature) in a manner which resembles training in machine learning. We can use a training set of examples for building the prompt, and a testing set for evaluating the LLM response.

To train the LLM via the DSPy signature we built a list of 198 examples using and readapting some of the content of the OA datasets cited above. The examples list mirrors the signature format and contains pairs of codes-descriptions and meaning. Some examples are shown in Table 1.

| *text_1 (**Code.** Description)* | *text_2 (**Code.** Description)* | *meaning* |
|---|---|---|
| **Reliability.** References to employment and (un)faithful application of (in)appropriate research methods and R&I related processes and procedures. | **Respect.** References to treatment of human participants with due consideration for their autonomy and dignity. | the two texts have a different meaning |
| **Medical responses- dismissive, unhelpful.** Expressing frustration at experiences where medical staff gave unhelpful advice/comments to the parent of did not believe them (where parent felt this) | **Medical problems of infant.** Describing medical issues that the infant had upon birth, during the NICU, and after discharge. Gives a sense of the vulnerability of premature infants and their common relationship with illness and the medical system | the two texts have a different meaning |
| **Other essential goods.** References and perceptions related to difficulties paying rent or bills, such as water and electricity during the economic recession | **Labour conditions.** References and perceptions related to changes in labour conditions in general during the economic recession | the two texts have a different meaning |
| **Self-doubt with parenting.** Expressing worries and doubts with thoughts of not meeting their child's needs or not knowing what to do | **Insecurity.** References related to concerns and uncertainties about fulfilling their child's requirements or the parents' lack of knowledge on how to handle the situation. | the two texts have a similar meaning |

---

[1] see https://dspy-docs.vercel.app/docs/building-blocks/signatures

| **Pursuing the truth.** References to attitudes and behaviours in relation to fabrication and falsification of data and plagiarism of research materials, data, and other information. | **Pursuing the truth.** References of respondents the mindsets and the actions related to the creation of false research data, to plagiarism and also the manipulation of information. | the two texts have a <u>similar</u> meaning |
|---|---|---|
| **Income loss**. References and perceptions related to loss of income during the economic recession. | **Financial setback**. How the respondents perceive and refer to the decline in their earnings and their income loss. | the two texts have a <u>similar</u> meaning |

Table 1 – Examples of pairs of codes for training – description, and meaning

The pairs with different meanings are simply different pairs of codes (and relative descriptions) from the same analysis (i.e. taken from the four OA analysis cited). For the pairs with similar meaning instead, we provided an LLM with the full list of text_1 used for the codes with different meaning (99 code-description pairs) and asked the model to provide a text with a similar meaning (both a code and a description). The text provided by the LLM was then manually modified and adjusted by us to make it look more similar in form to the text_1, i.e. making it look like it was written by a human rather than by the AI.

The DSPy pipeline then is as follows:

1. The 198 examples (99 different pairs and 99 similar pairs, and related meaning) are randomised and then split into a training set (80%) and a test set (20%).
2. The training set is used to train the LLM, with the DSPy module BootstrapFewShotWithRandomSearch, which generates demonstrations based on the training set, in this case demonstrations of the comparisons.
3. The 'RandomSearch' component applies a 'few shot'[2] prompting multiple times and then operates a selection of the best examples (based on validation scores, see point 5), which are then used for the prompt.
4. Then the results are 'compiled' with a ChainOfThought as predictor (i.e. when a new pair of text is provided, the ChainOfThought allows the LLM to predict if they are similar or different based on the Bootstrap and provides an explanation of how the model arrived at this conclusion).
5. DSPy also includes options for having validation metrics of the program. In this case we used a simple metric which compares the meaning in the testing set (i.e. either "the two texts have a similar meaning" or "the two texts have a different meaning") with the meaning of the prediction and returns True in case they are the same or False in case they are not.

---

[2] Few shot prompting is a prompting strategy where the model is shown a few examples of a task, before being asked to perform it.

The pipeline provides a compiled and validated prompt which includes a set of examples, and instructions for the LLM to perform the reduction of duplicate codes. Two examples (part of the full prompt) are shown in Box 3. The full prompt used for the research is available in the Appendix. The evaluation metric used within DSPy always returned True on the training set, showing its general applicability to the task at hand.

```
{
    "text_1": "Feelings of fear. Parent describes experiences in the NICU or post NICU of feeling scared in the moment or due to future possibilities ",
    "text_2": " Anxiety. A parent recounts their overwhelming experiences in the NICU or post NICU, where they were consumed by anxiety and uncertainties.",
    "meaning": "the two texts have a similar meaning"
},
{
    "text_1": "Linting tools. These codes are quality assurance and software engineering practices used to assure quality.",
    "text_2": "Accountability. Accountability is an expectation that an individual of the Scrum team will be evaluated on their performance including the quality of their work. It also means being answerable to the team's expectations.  ",
    "meaning": "the two texts have a different meaning"
}
```

Box 3 – Two of the examples included in the compiled DSPy prompt

The compiled prompt can be saved as a text file and used (without the need to compile the program every time) for all future codebook reduction processes (see Appendix).

For the codebook reduction, the compiled prompt is lastly passed again to a DSPy ChainOfThought module. Through this the LLM will see the chain of thought associated with the examples (i.e. it will see how pairs of text are deemed similar or different) and it will compare each code-description pair (for the interview sets) one by one with each code-description pair in the UCC. Once a code is deemed similar to another code in the UCC then the program moves on to compare the next, and the similar code is saved separately in a csv file. If the code instead does not match any of the unique codes, then it is added to the UCC. Figure 4 summarises visually the entire process.

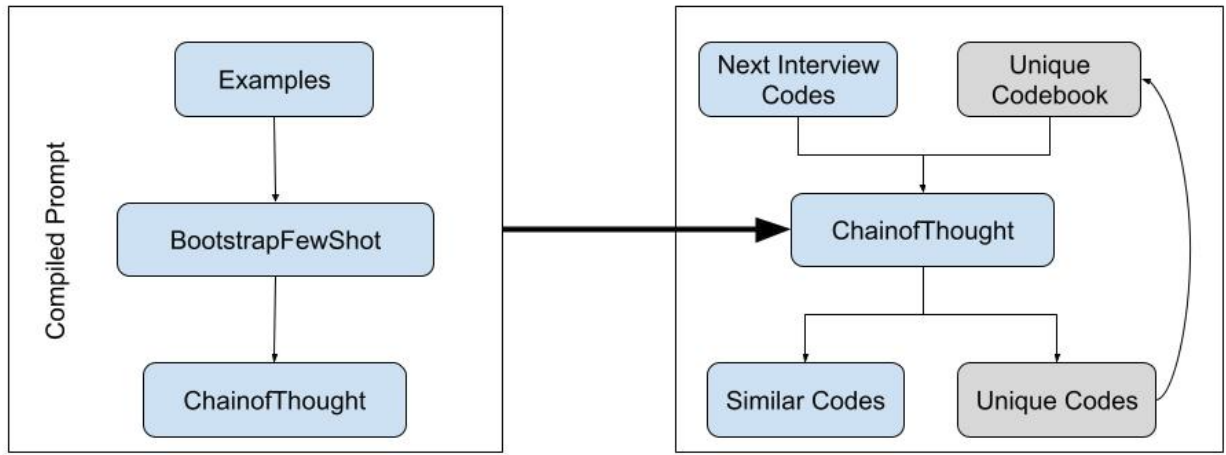

Figure 4: DSPy pipeline for the reduction of the codebook and identification of unique initial codes

**Material**

For performing this study, we propose to use a dataset of size and topic comparable to that of Costantinou et al. (2017), which conducted interview in a hospital setting. The dataset we use is titled *Interviews on Trials Transparency at Public Research Institutions in the UK, 2020-2021* (De Vito, 2023), which is part of a study that "aimed to collect the institutional perspective to transparency issues in clinical research." The interviews were conducted "with professionals working in research governance, trial management, and administrative leadership positions " in the UK National Health Service.

**Evaluation**

We propose an evaluation of our results. We will use semantic similarity scores to compare initial codes with calculation of cosine similarity scores for sets of unique codes (and related descriptions) obtained from the same process (e.g. Prompt_2 Vs. Prompt_2) but with codes from different iterations or sequences of analysis, and similarity scores of UCCs based on the same sequence and obtained with Prompt_2 or the DSPy prompt. The semantic similarity evaluation was done using the SBERT family of sentence transformer models. Each code was transformed into a high-dimensional vector, or embedding, capturing its semantic meaning. Cosine similarity, the difference in angles between these embedded code vectors, was then computed to quantify similarity. For the code-level similarity, we used the all-MiniLM-L6-v2 – a small but efficient sentence transformer suitable for tasks requiring comparisons across a long list of sentence pairs.

# Results

## Initial coding

The output from an LLM is inherently probabilistic due to its fundamental design. At the core, LLMs use probabilistic algorithms based on statistical patterns in the training data. Additionally, the hardware and software architectures, such as GPUs and random initialization of parameters, introduce elements of non-determinism. Consequently, given the same input, the model may produce slightly different outputs across different runs, even if it is operated with Temperature at 0 (that is with 'limited or no creativity)'. As we can see in Table 2, *with the same sequence of interviews* (from 1 – 12) on 7 iterations/runs using Prompt_1, the total number of initial codes produced differs from iteration to iteration, ranging from a minimum of 162 to a maximum of 182. This occurs even if in Prompt_1 we explicitly asked the model for 15 codes (up to) for each interview in the dataset. We can observe therefore variation in the LLM response, e.g. the 20-code gap between iteration *3 and iteration *5. This gap may have implications for the analysis and calculation of the ITS saturation metric.

| Iteration | Nr. Of Initial Codes |
|---|---|
| *1 | 175 |
| *2 | 167 |
| *3 | 162 |
| *4 | 176 |
| *5 | 182 |
| *6 | 168 |
| *7 | 170 |

Table 1: Total number of initial codes from the LLM, with Prompt_1, same sequence (1 to 12) on different iterations

| Sequence | Nr. Of Initial Codes |
|---|---|
| #1 | 176 |
| #2 | 169 |
| #3 | 187 |
| #4 | 163 |
| #5 | 168 |
| #6 | 177 |

Table 2: Total number initial codes from the LLM, with Prompt_1, different sequences

Table 2 shows the initial codes from the LLM when *different sequences* of the interviews are used. The first 4 are the same sequences used by Costantinou et al. (2017) (albeit we should remind we are using a different set of interviews), #5 and #6 are two additional sequences. There is variation again in the codes produced by the LLM, however it does seem difficult to attribute this according to what Costantinou et al. (2017) called the order-induced error (albeit here interviews are coded independently from one another). As the figures in Table 2 appear comparable to those in Table 1, except perhaps for #3. It is more likely that the variations in Table 2 (like those in Table 1) are attributable to the probabilistic response of the LLM, than to the order in which interviews are analysed. Although we cannot a-priori exclude some effects related to the order. Table 3 shows, for completeness, some examples of codes obtained using Prompt_1.

| Code | Description | Quote |
| --- | --- | --- |
| Reasons for Trial Registration and Reporting | The participant's views on the importance of trial registration and reporting for transparency and good governance. | I think the reason that those are in place it's a number of reasons; transparency first and foremost so that it's publicly accessible to see what's running, how it ran allowing the public to review any ongoing trials potentially give them access to trials they may be interested in and I think general again really good governance oversight so that's a reportable structure. |
| Disagreements and Downsides | The potential disagreements or downsides that can arise from trial registration and reporting requirements. | Generally, people not having study teams not having the time to update these systems, forgetting about them, not really understanding the value of them was a problem I believe. |
| Process of Trial Registration | The participant's involvement in the process of trial registration at their institution. | So, at the moment it is very much about myself being proactive in contacting researchers, so I'm aware of everything that's coming up. |

Table 3: Examples of codes from the LLM

**Reduction of the Codebook: from TCC to UCC**

The initial coding procedure is based on performing the analysis of each interview separately and independently. We know this leads to duplications in the codebook. As we discussed, we can perform a removal of duplicates and, with the results of this process, also measure saturation using a metric we call ITS. Table 4 presents again the total number of initial codes of Table 2 as well as the total number of unique codes (without duplicates) obtained with the Prompt_2 ('old' procedure') and the number obtained with the DSPy procedure. ITS is computed by dividing the total number of unique codes obtained from each method divided by the total number of initial codes (or number of 'total unique codebook').

| Iteration | Total Nr. Of Initial Codes | Total Nr. Of Unique Codes (Prompt_2) | ITS (Prompt_2) | Total Nr. Of Unique Codes (DSPy) | ITS (DSPy) |
|---|---|---|---|---|---|
| *1 | 175 | 71 | 0.41 | 70 | 0.40 |
| *2 | 167 | 50 | 0.30 | 71 | 0.43 |
| *3 | 162 | 45 | 0.28 | 70 | 0.43 |
| *4 | 176 | 42 | 0.24 | 74 | 0.42 |
| *5 | 182 | 73 | 0.40 | 72 | 0.40 |
| *6 | 168 | 65 | 0.39 | 71 | 0.42 |
| *7 | 170 | 80 | 0.47 | 66 | 0.39 |

Table 4: Unique codes and ITS with Prompt_2 and the DSPy process, iterations of same sequence of interviews (from 1 to 12)

We can observe that the total number of unique codes (after the reduction) is consistent when DSPy is used. But, when the reduction is done with Prompt_2, there is more variation. The ITS metric using the DSPy procedure is highly consistent, with a range of only 0.04 between the largest and smallest values (Coefficient of Variation - CoV at 3.88%), The ITS appears more dispersed when Prompt_2 is used for the reduction (CoV at 23.34%)

Similar observations apply when looking at the results of the reduction with the initial codes from the different sequences (see Table 5). Using Prompt_2 we see variations in the total number of unique codes, whereas the DSPy prompt produces a fairly consistent reduction. Even in the case of #3, where the number of total codes is 24 codes higher than #4, we can see consistency if we look at the ITS metric. The higher number of codes (81 in #3) may be explained by considering the starting initial codes are 187, higher than all the other sequences. Also, for the sequences, the ITS metric value shows minimal variation when using the DSPy prompt, with a range of circa 0.03 (CoV at 2.79%) and it appears more dispersed when Prompt_2 is used (CoV 19.4%).

| Sequence | Total Nr. Of Initial Codes | Total Nr. Of Unique Codes (Prompt_2) | ITS (prompt_2) | Total Nr. Of Unique Codes (DSPy) | ITS (DSPy) |
|---|---|---|---|---|---|
| #1 | 166 | 57 | 0.34 | 70 | 0.42 |
| #2 | 169 | 48 | 0.28 | 70 | 0.41 |
| #3 | 187 | 66 | 0.35 | 81 | 0.43 |
| #4 | 163 | 40 | 0.25 | 65 | 0.40 |
| #5 | 168 | 71 | 0.42 | 72 | 0.43 |
| #6 | 177 | 71 | 0.40 | 74 | 0.42 |

Table 5: Unique codes and ITS with prompt_2 and the DSPy process with difference sequences

Figure 5 proposes a plot of the cumulative codebooks (both TCC and UCC) for Iteration *1 (which was selected here because of the similar number of unique codes resulting from the two reduction procedures). We see the TCC (in green) growing much faster than the UCCs. It is on the basis of this observation that the ITS metric was defined in our previous work, as we discussed in the methods section of the paper. The UCCs from Prompt_2 and the DSPy prompt, we suggest, show a fairly comparable distribution instead. The regression curves for these (see Figure 6) are as follows Y_DSPy = 4.68 * X + 21.44 (DSPy) and Y_Prompt_2 = 4.47 * X + 20.99 (Prompt_2), with a Mean Square Error of 3.1 for the differences between the two. Later in the evaluation of the results we will discuss them in more detail aspects of this similarity, from a content perspective.

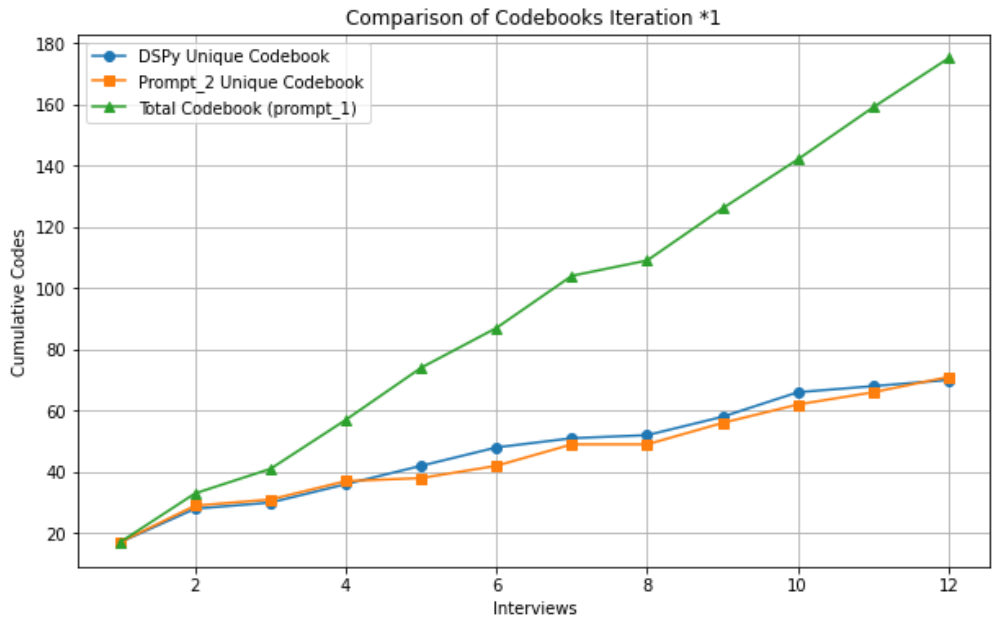

Figure 5: TCC (Prompt_1) and UCCs Prompt_2 and DSPy in Iteration *1

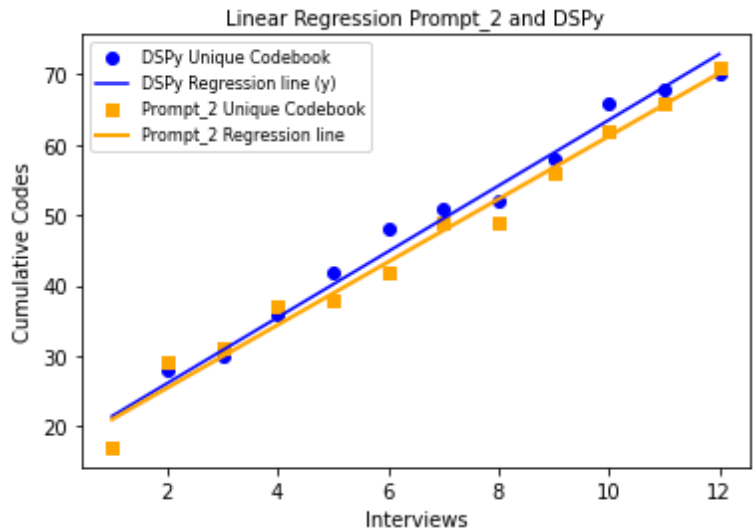

Figure 6: Regression lines comparison for Iteration *1 (DSPy & Prompt_2), UCCs

Figure 7 shows the UCCs of Iteration *3, for Prompt_2 and the DSPy prompt. This iteration was chosen for display because of the gap in total unique codes (45 Vs 70) between the two. We can see how they diverge from the first comparison, ending with two codebooks of different sizes. The ITS in this case differs substantially between the two as seen in Table 2 for *3.

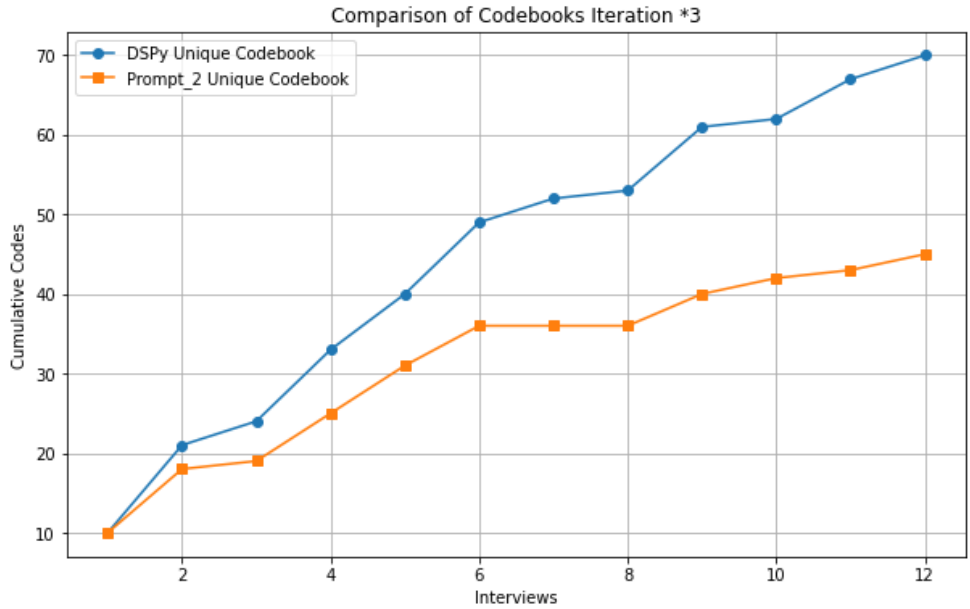

Figure 7: Unique codes from Prompt_2 and DSPy in Iteration *3

Figures 8 shows a further comparison of UCCs, in these cases mixing different Sequences/Iterations, which present the same or a very close number of unique codes (#5 and #6 using the DSPy procedure). In Figure 9 the two linear regressions of these set of codes are presented, showing a close and comparable curve. For example, Figure 8 shows the unique codes obtained with the DSPy prompt for the sequences #5 and #6. All in all, whilst there are some localised differences in the shape of the figures (also considering in these cases we are starting from a different initial coding), these differences are not enormous, when the UCCs are of comparable size. Differences however exist when the UCCs are fairly different in size (as in Figure 7).

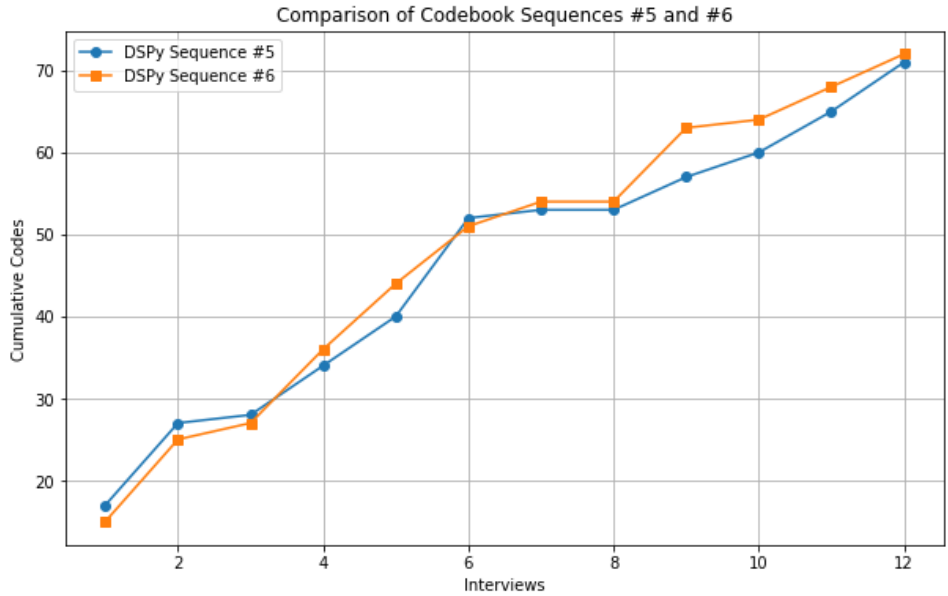

Figure 8: Unique codes from Sequence #5 and #6 only using the DSPy prompt

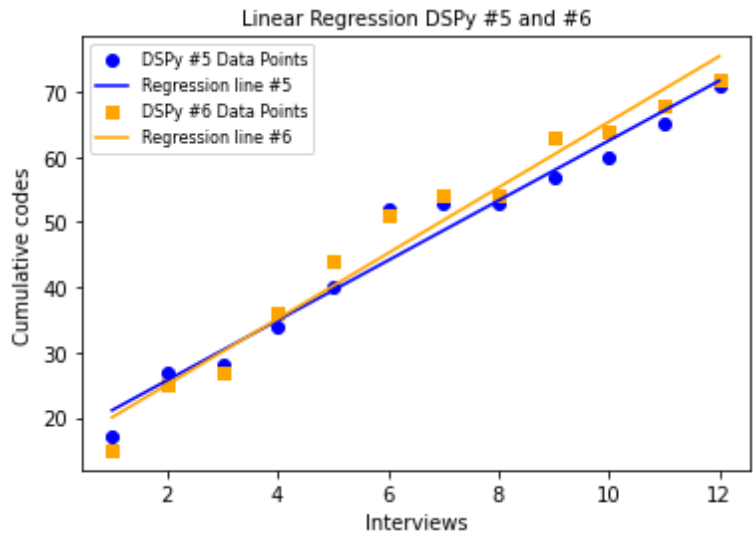

Figure 9: Regression, sequence #5 and #6 only using the DSPy promp

Lastly Figure 10, puts several UCCs in the same cartesian axis. This allows us to appreciate, that whilst there are some differences in the shape of each codebook, the shapes appear quite comparable overall.

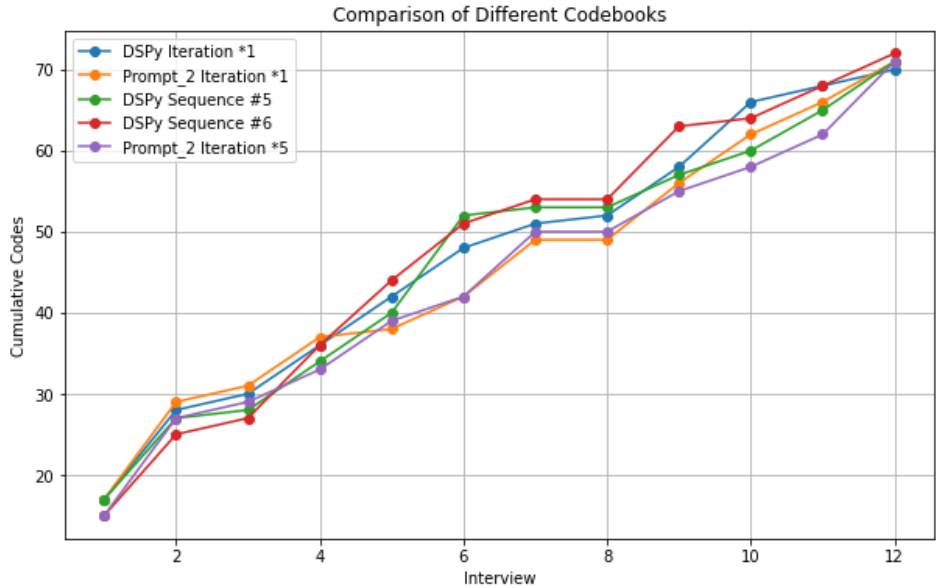

Figure 10: Wider comparison of 5 UCCs of similar size (from Prompt_2 and DSPy prompt)

**Evaluation**

For evaluating the results first, we propose to look at Figure 11, Figure 12 and 13. These Figures show the results of semantic similarity scores – using vector embedding and cosine similarity scoring – between two different UCCs. Boxes in red are between code pairs that are found to be semantically similar. Figure 11 shows the similarity for the unique codes of Iteration *1 (with Prompt_2 and DSPy). In the diagonal we see the similarity of codes. The first 17 codes are the same (in both procedures they amount to the codes for the first interview) and we can see the perfect diagonal in red. Nonetheless we can clearly observe around the diagonal the red squares with the similarity between pairs of codes approximating to one. Note that the codes were reordered (using the Kuhn-Munkres algorithm from the python scipy library) in the matrix to display maximum similarity in the diagonal. Overall, the unique codebooks for Prompt_2 and the DSPy prompt, derived from the same set of initial codes (*1), do appear comparable.

Figure 12 shows the semantic similarity for the unique codes for Iteration *1 and *3 only using the DSPy prompt. Observations were again reordered for maximum similarity. We can still perceive from the figure the diagonal in red (albeit this looks less marked), which tells us the codebooks may be comparable in terms of semantic likeness. As expected for the first sequences of codes the similarity is not as strong as in Figure 11. This is because in Figure 11 we are comparing unique codebooks derived from different initial coding processes, albeit on the same data. Lastly Figure 13 shows the semantic similarity for codes derived with the DSPy prompt for Iteration *3 and Sequence #5. The diagonal in red is again clearly visible, suggesting the two codebooks do present some degree of semantic similarity.

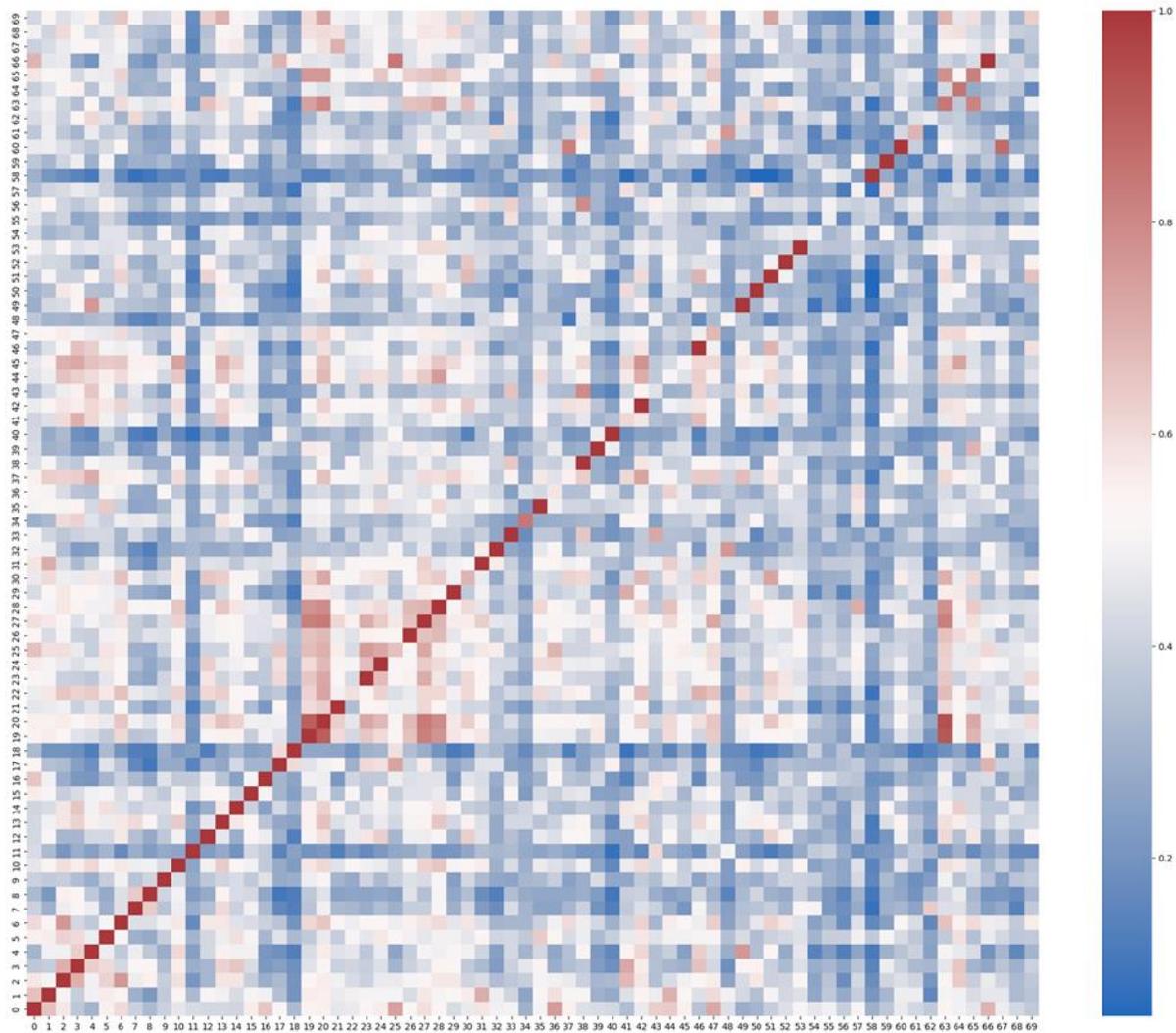

Figure 11: Semantic similarity matrix for unique codes in Iteration *1 (Prompt_2 vertical, DSPy horizontal

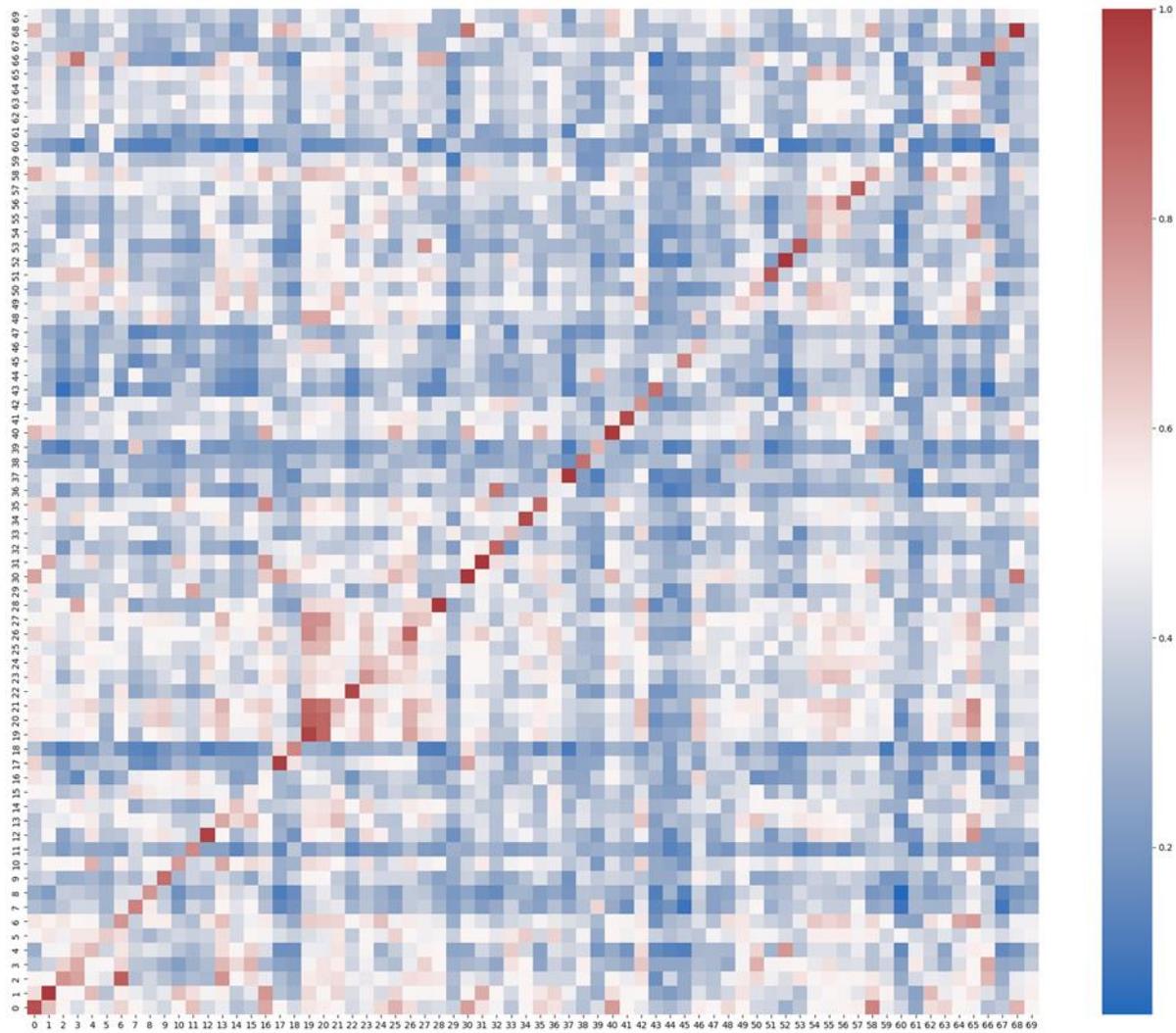

Figure 12: Semantic similarity matrix for unique codes in Iteration *1 and *3 using the DSPy prompt

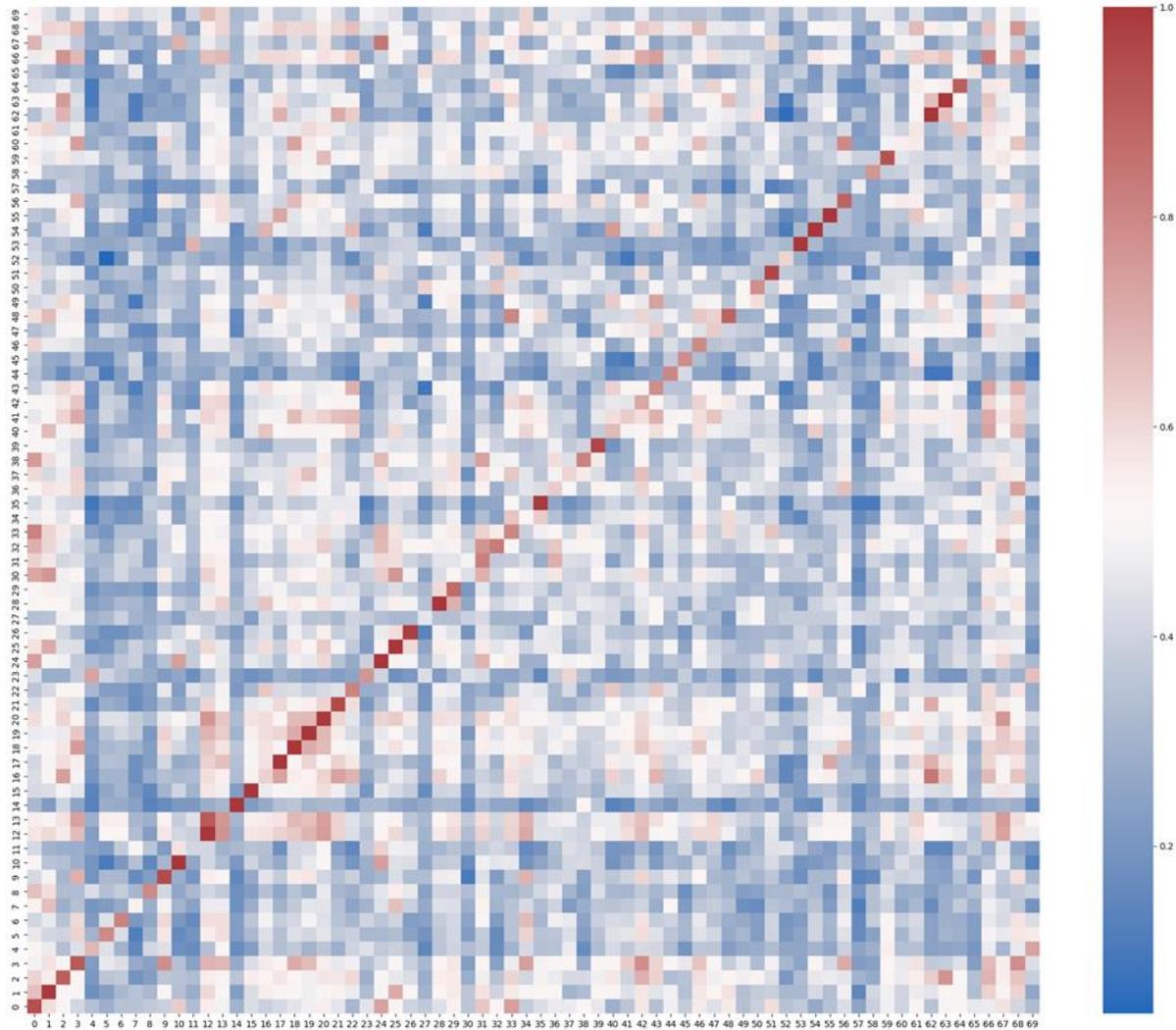

Figure 13: Semantic similarity matrix for unique codes in Iteration *3 and Sequence #5 using the DSPy prompt

Although there are some gaps in the semantic similarity, as seen in the examples of Figure 11-13, we can assume that the codebooks may generally be comparable and similar, as the diagonal in the figures shows clearly the presence of similarity. Some differences are to be expected for a variety of reasons, such as the probabilistic response of LLMs to prompts, and the fact that the codebook reduction is also performed on different initial coding, including different sequences, as in the cases #2 to #6 and with different procedures. In fact, if we look at Figure 13, where the sequences are different, it is rather encouraging to see a clearly defined diagonal with some strong codes semantic similarity.

## Discussion and Conclusion

The concept of saturation has been extensively debated in qualitative research, with scholars emphasising its manifold nature (Saunders et al., 2018; Sebele-Mpofu, 2020), as well sometimes its problematic application (O'Reilly and Parker, 2012; Braun and Clarke, 2021). When qualitative analysis is performed with LLMs some of these observations need to be addressed. In our work, we focused on Inductive Thematic Saturation (ITS) within the context of using LLMs for Thematic Analysis. Our findings align and extend existing literature, particularly regarding the analytical properties of saturation, as well as the use of metrics to assess saturation. Since LLMs are not involved in decisions about data collection or sampling, analytical saturation is key over other types of saturation. Moreover, the works of Lowe et al. (2018) and Guest et al. (2020) introduced metrics to measure saturation, emphasising aspects related to codes generation and diminishing returns. Our study complements these efforts by proposing an approach to measure saturation in LLM-driven TA.

This paper's overall aim was to develop further observations on measuring saturation for initial coding and Thematic Analysis performed with LLMs. The paper had two objectives: a primary objective related to proposing a novel technique for deriving a final codebook of initial codes (the Unique Cumulative Codebook), predicated on a secondary objective related to assessing some of the emergent behaviour of LLMs when performing initial coding.

We will start reflecting on the secondary objective. We have observed how an LLM's initial coding (performed with Prompt_1) may produce a different number of initial codes on repeating the process over several iterations, even when a model has Temperature at zero (i.e. 'no creativity'). This behaviour we attributed to the probabilistic nature of the LLM response. In other words, there can always be some variation in the model response to the same task. Clearly this behaviour may have some implications for an overall analysis and also for the way we proposed to measure saturation, via the ITS metric. We performed also the initial coding on different sequences of interviews also, to assess whether there is an order-induced error, a concept originally proposed by Costantinou et al. (2017) for human initial coding. However, comparing the number of initial codes for the different sequences we used, or the repetition of the initial coding on the same sequence, it is not possible to measure the presence of any such error. We need to remind however that differently from Costantinou et al. (2017), the initial coding with LLMs is done on each interview independently. In any case, what can be observed is in fact only the variation in the response, with the model producing different numbers of initial codes. Whilst we cannot exclude a-priori the presence of an order-induced error, it is difficult to prove its presence, given the variation in codes a model produces on iterations of the same sequence.

Following we performed the reduction of the codebook from duplicates, to arrive at a set of unique codes (without repetitions, the UCC). As we explained in previous pages, this is a step currently necessary when doing TA with LLMs, as the interviews are coded independently, one at a time with the LLM having no knowledge of the previous codes or interviews. In a Total Cumulative

Codebook there are therefore duplicate codes or codes that are very similar to one another, and the reduction task aims at finding these duplicates and removing them to arrive at a UCC that characterise the dataset. It is then the ratio between the overall number of unique codes and the overall number of total codes that gives the ITS saturation metric. However, since we observed that the initial coding may produce different sets of total codes when performed on multiple iterations or diverse sequences, this may have implications for measuring the ITS metric. In other words, the variation in the LLM response impacts the measure of the ITS metric, however it may be possible to explore solutions for a more predictable measure of this form of saturation.

To address this problem, we performed the reduction of the codebook with two procedures. The first one used a zero-shot 'traditional prompt' (the same we used in our previous works, i.e. Prompt_2), the second instead used a novel procedure based on the DSPy framework – where machine learning approaches are used to optimize prompts toward desired functionality. We observed in the case of the 'traditional prompt' variation in the number of unique codes obtained as response, with clear variation on the saturation metric. This variation may also be dependent on the TCC. When using the DSPy prompt, the variation appears marginal, and the ITS metric more consistent (with only very small differences, as measured by a CoV). Overall, we would suggest that using the DSPy procedure is desirable as it delivers consistent analytical saturation, and a more precise measure of the ITS metric. Nonetheless, this does not mean that the 'traditional prompt' process is not valid, as will be commented below.

We have performed an evaluation on several levels. We performed a semantic similarity comparison between some example of the UCCs obtained from the 'traditional prompt' and the DSPy prompt, when the number of codes is comparable in number (e.g. in iteration *1), the results show a good degree of similarity on the comparison matrix diagonal, which suggests that both procedures are actually equivalent when the 'traditional prompt' process saturates at the same level as the DSPy prompt. This suggests that the Prompt_2 process is actually a viable solution for measuring saturation, when the reduction performs consistently. However, because of the variations observed over multiple repetitions, there is always some risk that the saturation is not measured correctly.

The results of this work confirmed the validity of the ITS metric we proposed in our previous work, considered as the ratio of the overall unique and total codes. Whilst there is variation in the Prompt_2 results, the reduction procedure with DSPy shows strong consistency in the ITS metric. Thus, any issue with this measure as caused by 'imprecise' procedures (e.g. Prompt_2) may be entirely mitigated.

**Implications for practice**

There are potentially several implications for the practice of performing TA with LLMs, stemming from this work.

First, analysts can decide whether to use the 'traditional prompting' or the DSPy prompting. As we have seen they can be equivalent. But in the case of Prompt_2 it would be advisable to repeat the initial coding process multiple times, to assess which is the potentially correct number of initial codes. The DSPy procedure does seem to possess the advantage of offering consistency across the board, but it is more complex from a technical point of view (i.e. the python code is much more complex) and it potentially is also more expensive to run. However, the DSPy procedure carries a heavier overhead in terms of use.

The reduction process we believe is more expensive (e.g. if a paid LLM) is used. Moreover, programming an LLM via DSPy requires far more advanced programming skills than using the traditional procedure. Notwithstanding this, the DSPy prompt we used is available and can be reused by other researchers for their analysis and reduction of the codebook. Thus, potentially researchers would not be required to compile the prompt with DSPy, but just use it for their work (i.e. only perform the process represented in the right side of Figure 4).

A further implication is the recommendation about repeating the analysis process more than one time to confirm the results. As we observed there is variation in the response of a LLM, especially in the initial coding phase.

Further testing will be necessary to improve the new procedure for the codebook reduction. For compiling the DSPy prompt we used four existing exemplar datasets, additional datasets may be used to assess if more variation in the training material for the prompt generates diverse results. Nonetheless what we used and proposed here shows how the new procedure allows for consistency in measuring the ITS for LLMs.

**APPENDIX (3500 words)**

1. **Dspy prompt for reduction of the codebook**

```
{
 "generate_answer": {
  "lm": null,
  "traces": [],
  "train": [],
  "demos": [
   {
    "augmented": true,
    "text_1": "Open access. References to making R&I results publicly available (e.g. through open access publications) and \/ or to arguments for acting otherwise.",
```

```
    "text_2": "Engaging with decision-makers. References to awareness of, and as far as possible, avoidance of actual or perceived conflicts of interest of the researchers and\/or organisations performing the R&I.",
    "rationale": "produce the meaning. We will analyze the content of Text 1 and Text 2 and compare them to determine if they have a similar or different meaning.",
    "meaning": "the two texts have a different meaning"
   },
   {
    "augmented": true,
    "text_1": "Medicines and health services. References and perceptions related to difficulties buying medicines and accessing health services during the economic recession",
    "text_2": "Other essential goods.  References and perceptions related to difficulties paying rent or bills, such as water and electricity during the economic recession",
    "rationale": "produce the meaning. We will compare the references and perceptions related to difficulties buying medicines and accessing health services (Text 1) with the references and perceptions related to difficulties paying rent or bills, such as water and electricity (Text 2). By comparing these two texts, we can determine if they have the same meaning or a different meaning.",
    "meaning": "the two texts have a different meaning"
   },
   {
    "augmented": true,
    "text_1": "Organizations. References to regulation of R&I ethics on organizational level",
    "text_2": "Research Organisations. Aspects related with the ethics and governance of research and innovation at the organisational level.",
    "rationale": "produce the meaning. We can see that both texts are discussing the regulation and governance of research and innovation ethics at the organizational level. While the wording may be slightly different, the overall meaning is the same.",
    "meaning": "the two texts have a similar meaning"
   },
   {
    "augmented": true,
    "text_1": "Accountability. Accountability is an expectation that an individual of the Scrum team will be evaluated on their performance including the quality of their work. It also means being answerable to the team's expectations.  ",
    "text_2": "Linting tools. These codes are quality assurance and software engineering practices used to assure quality.",
    "rationale": "produce the meaning. We will analyze the definitions and descriptions of the two texts and compare them to determine if they have a similar or different meaning.",
    "meaning": "the two texts have a different meaning"
   },
   {
    "text_1": "Worries & fears with developmental milestones . Expressing fears for future regarding infant meeting developmental milestones and current fears of infant not meeting milestones presently ",
```

```
    "text_2": "Difficulty leaving baby. Describing leaving the baby in the NICU and then when at home as hard and emotionally stressful. ",
    "meaning": "the two texts have a different meaning"
  },
  {
    "text_1": "Emotional needs\/bond (not feeling like a parent) . Describes memories where the parent feels they were not able to engage in the typical emotional care needs of their child (being present, looking at baby, holding them) or bond with them. ",
    "text_2": "Parental Disconnect. Observations where the parents mention cases where they struggled to fulfill their child's needs and establish a bond. For example making eye contact.",
    "meaning": "the two texts have a similar meaning"
  },
  {
    "text_1": "Worries & fears with developmental milestones . Expressing fears for future regarding infant meeting developmental milestones and current fears of infant not meeting milestones presently ",
    "text_2": "Concerns about developmental milestones. References to concerns about the future regarding the ability of the baby to reach developmental milestones and worries about the progress.",
    "meaning": "the two texts have a similar meaning"
  },
  {
    "text_1": "Software quality is a complex concept. Two participant (out of eight) stated that software quality is a 'complex' concept, i.e., made up of different attributes that are linked complicated way.",
    "text_2": "Understanding Software Quality. Two individuals (amongst eight) said that software quality is a difficult notion because it connects with many different things.",
    "meaning": "the two texts have a similar meaning"
  },
  {
    "text_1": "Individuals. References to self-regulation practices of R&I ethics on individual level.",
    "text_2": "Individual autonomy. References to obtaining informed and voluntary consent from human participants (or their legal guardians) for taking part in R&I related activities and provide personal data and other information.",
    "meaning": "the two texts have a different meaning"
  },
  {
    "text_1": "External quality. This code was used to categorize some software quality attributes (the above columns), e.g., free of defects and conformity to business needs).",
    "text_2": "External quality. This script was employed to classify various attributes of software quality toward the external world, such as absence of flaws and alignment with business requirements.",
    "meaning": "the two texts have a similar meaning"
  },
  {
```

```
    "text_1": "Income loss. References and perceptions related to loss of income during the economic recession",
    "text_2": "Financial setback. How the respondents perceive and refer to the decline in their earnings and their income loss.",
    "meaning": "the two texts have a similar meaning"
},
{
    "text_1": "Empowerment and Self-reflection . This node summarizes why a person with dementia may be involved in the end-user study, speaking to how their involvement in this type of research creates a sense of empowerment for the participant.",
    "text_2": "Promoting Independence. Observations about including individuals with dementia in the end-user study, this includes how this fosters empowerment for the participant.",
    "meaning": "the two texts have a similar meaning"
},
{
    "text_1": "External quality. This code was used to categorize some software quality attributes (the above columns), e.g., free of defects and conformity to business needs).",
    "text_2": "Early engagement of QAs. This is another facet of the collaborative aspect of Scrum. The engagement of QA team members early on the development process shows that Scrum teams proactively engage rather than wait and see. Compared to the phased approach of software development where the QAs wait until the code is ready for testing, in Scrum, all team members are engaged early on the process to assume their responsibilities and contribute collaboratively in achieving the desired outcome.",
    "meaning": "the two texts have a different meaning"
},
{
    "text_1": "Emotional needs\/bond (not feeling like a parent) . Describes memories where the parent feels they were not able to engage in the typical emotional care needs of their child (being present, looking at baby, holding them) or bond with them. ",
    "text_2": "Trauma and shock.  Where parent describes situations where they felt it was traumatic or created a direct shock response (parent will use these words or use them indirectly through examples) ",
    "meaning": "the two texts have a different meaning"
},
{
    "text_1": "Ambiance\/Physical Space . This node describes how participants felt the ambiance, e.g., a relaxed interview atmosphere with no feelings of pressure on them, and the physical space of the user experience research procedure itself, made them feel comfortable. This 'comfort' allowed participants to fully engage in the user experience research process without feeling shame\/embarrassment for not knowing something, asking questions, etc. Participants also felt free to criticize the app. This atmosphere facilitated truthful participation.",
    "text_2": "Pace. This node summarizes an important point of feedback from participants, namely that the interviewer conducted the user experience research process at a proper pace, which is important given the cognitive impairment exhibited by people with dementia.",
    "meaning": "the two texts have a different meaning"
},
```

```
    },
    {
      "text_1": "Medical\/psychological problems of parent . Descriptions provided of physical problems that the parent had either before, during, after the birth and also psychiatric problems the parent mentions having or believing to have had. ",
      "text_2": "Sense of responsibility for health of baby.  Expressing thoughts and emotions related to feeling responsible for the prematurity ",
      "meaning": "the two texts have a different meaning"
    },
    {
      "text_1": "User Experience Research Location. Comments about how app user experience research sessions could be done by phone, problems with Zoom, etc.",
      "text_2": "Remote User Research. References from participants about alternative methods for conducting user experience research remotely, such as conducting sessions over  Zoom and other solutions.",
      "meaning": "the two texts have a similar meaning"
    },
    {
      "text_1": "Self-doubt with parenting . Expressing worries and doubts with thoughts of not meeting their child's needs or not knowing what to do ",
      "text_2": "Insecurity. References related to concerns and uncertainties about fulfilling their child's requirements or the parents' lack of knowledge on how to handle the situation.",
      "meaning": "the two texts have a similar meaning"
    },
    {
      "text_1": "Quality assurance. The set of practices, techniques and processes used to assure the quality requirements of the software. This code emerged for the data because it is part of one causal chain discussed by Participant 1 (see Causal Chains RQ1 tab). Other codes (e.g., Assuring clean code) were used to infer quality assurance.",
      "text_2": "Peer review. These are quality assurance practices.",
      "meaning": "the two texts have a different meaning"
    },
    {
      "text_1": "knowledge sharing. Our participants accounts show that knowledge sharing is the process of integrating individually held information and know-how into the common knowledge of the development team. This takes the form of either peer to peer learning or business user (sometimes through the proxy of the PO role) to developers (including QAs). Peer to peer learning is a practice in which Scrum team members exchange information and help each other's with their knowledge during the interaction with one another.  This can happen during the Stand Ups or simply when a developers walk to his colleague to ask questions. Business users to developers knowledge sharing mainly concerns a better understanding the requirements.",
      "text_2": "Sharing Knowledge. The participants said the sharing knowledge creates some form of collective knowledge of the development team. This can occur via peer-to-peer learning or through communication. ",
      "meaning": "the two texts have a similar meaning"
    }
```

```
    },
    {
      "text_1": "Returning to normality . Parent describes milestones or feelings where they feel that things are returning to how they were meant to be and back to a normative parent/family experience ",
      "text_2": "Lost experiences.  Expressing missing out on an experience of a full-term birth and what the prematurity meant for the parents' identity and preformed ideals of what they would be doing as a parent of a new-born baby. ",
      "meaning": "the two texts have a different meaning"
    },
    {
      "text_1": "Precarious work conditions. References and perceptions related to precarious work conditions, such as pressure, stress and insecure work during the economic recession",
      "text_2": "Precarious work conditions. References related with with challenging work environments, like difficult demands, tension with employers and uncertain employment due to the financial crisis.",
      "meaning": "the two texts have a similar meaning"
    },
    {
      "text_1": "Respect. References to treatment of human participants with due consideration for their autonomy and dignity.",
      "text_2": "Reliability. References to employment and (un)faithful application of (in)appropriate research methods and R&I related processes and procedures. ",
      "meaning": "the two texts have a different meaning"
    },
    {
      "text_1": "Early engagement of QAs. This is another facet of the collaborative aspect of Scrum. The engagement of QA team members early on the development process shows that Scrum teams proactively engage rather than wait and see. Compared to the phased approach of software development where the QAs wait until the code is ready for testing, in Scrum, all team members are engaged early on the process to assume their responsibilities and contribute collaboratively in achieving the desired outcome.",
      "text_2": "Involvement of QA Team. References to the idea that QA team members are involved from the beginning of the development process. This is different from traditional software development.",
      "meaning": "the two texts have a similar meaning"
    },
    {
      "text_1": "Psychological safety. Our participants used, for example, less feared and no "cconsequence", to make reference to a psychological safe working environment. Our participants explained that promoting a sense of psychological safety empowers all involved people to feel comfortable showing initiative, investing effort on achieving quality and caring about the quality of the deliverables.",
      "text_2": "Conformity to business needs.  These codes are the attributes that constitute software quality. In some instances participants used different terms to mean the same thing. For
```

example, \u0093meet business needs' and 'Meet the requirements of the end user' were used to imply conformity to business needs.",
      "meaning": "the two texts have a different meaning"
    },
    {
      "text_1": "Feelings of fear . Parent describes experiences in the NICU or post NICU of feeling scared in the moment or due to future possibilities ",
      "text_2": " Anxiety. A parent recounts their overwhelming experiences in the NICU or post NICU, where they were consumed by anxiety and uncertainties.",
      "meaning": "the two texts have a similar meaning"
    },
    {
      "text_1": "Linting tools. These codes are quality assurance and software engineering practices used to assure quality.",
      "text_2": "Accountability. Accountability is an expectation that an individual of the Scrum team will be evaluated on their performance including the quality of their work. It also means being answerable to the team's expectations.  ",
      "meaning": "the two texts have a different meaning"
    },
    {
      "text_1": "Software quality. This code is the topic of RQ1. Some participants used 'product quality' instead of software quality. This code emerged for the data because it is part of multiple causal chains ",
      "text_2": "Quality of Software. A few participants referred to it as \"product quality\" instead of software quality.",
      "meaning": "the two texts have a similar meaning"
    },
    {
      "text_1": "Continuous improvements. This practice is advocated by the agile manifesto (Principle 12). It calls for the agile team to continuously identify impediments and acted upon them. Other codes (e.g., Reflecting on the Scrum process and Continuous reflections) used by participant to imply continuous improvements. There also codes (i.e., Learning, Continuous experimentation and adapting) which are part of the continuous improvements process.",
      "text_2": "Constinuous software improvements. This is related to the agile manifesto (Principle 12) and the agile team should consistently recognise and address obstacles. Respondents use various methods for this (e.g., Analysing the Scrum process and Ongoing introspection) to signify ongoing enhancements.",
      "meaning": "the two texts have a similar meaning"
    },
    {
      "text_1": "Medical responses- dismissive, unhelpful . Expressing frustration at experiences where medical staff gave unhelpful advice\/comments to the parent of did not believe them (where parent felt this) ",

```
      "text_2": "Medical problems of infant.  Describing medical issues that the infant had upon birth, during the NICU, and after discharge. Gives a sense of the vulnerability of premature infants and their common relationship with illness and the medical system ",
      "meaning": "the two texts have a different meaning"
    }
  ],
  "signature_instructions": "Given the fields `text_1`, `text_2`, produce the fields `meaning`.",
  "signature_prefix": "Meaning:",
  "extended_signature_instructions": "Given the fields `text_1`, `text_2`, produce the fields `meaning`.",
  "extended_signature_prefix": "Meaning:"
 }
}
```